\documentclass{article}

\usepackage{microtype}
\usepackage{graphicx}
\usepackage{booktabs} 
\usepackage{caption}
\usepackage{subcaption}
\usepackage{kotex}
\usepackage{amsmath}
\usepackage{algorithm}
\usepackage{algorithmic}
\usepackage{color}
\usepackage{amssymb}

\newcommand{\nj}[1]{\textcolor{black}{#1}}
\newcommand{\sh}[1]{\textcolor{black}{#1}}
\newcommand{\ky}[1]{\textcolor{black}{#1}}

\usepackage{hyperref}

\usepackage[accepted]{icml2021}

\icmltitlerunning{Edge Bias in Federated Learning and its Solution by Buffered Knowledge Distillation}

\begin{document}

\twocolumn[
\icmltitle{\nj{Edge Bias in Federated Learning and its Solution \\ by Buffered Knowledge Distillation}} 



\icmlsetsymbol{equal}{*}

\begin{icmlauthorlist}
\icmlauthor{Sangho Lee}{snu}
\icmlauthor{Kiyoon Yoo}{snu}
\icmlauthor{Nojun Kwak}{snu}
\end{icmlauthorlist}

\icmlaffiliation{snu}{Graduate School of Convergence Science and Technology,
Seoul National University, Seoul, South Korea}

\icmlcorrespondingauthor{Sangho Lee}{shlee223@snu.ac.kr}
\icmlcorrespondingauthor{Nojun Kwak}{nojunk@snu.ac.kr}

\icmlkeywords{Machine Learning, ICML}

\vskip 0.3in
]



\printAffiliationsAndNotice{}  

\begin{abstract}
Federated learning (FL), which utilizes communication between the server (core) and local devices (edges) to indirectly \nj{learn from} more data, \nj{is an emerging field in deep learning research}. 
Recently, \textit{Knowledge Distillation-based} FL methods with notable performance and high applicability have been suggested. In this paper, we choose knowledge distillation-based FL method as our baseline and tackle a challenging problem that ensues from using these methods. \nj{Especially, we focus on the problem incurred in the server model that tries to mimic different datasets, each of which is unique to an individual edge device.} 
We dub the problem `\textit{edge bias}’, which occurs when multiple teacher models trained on different datasets are used individually to distill knowledge. We introduce this nuisance that occurs in certain scenarios of FL, and to alleviate it, we propose a simple yet effective distillation scheme named `\textit{buffered distillation}’. In addition, we also experimentally show that this scheme is effective in mitigating the straggler problem caused by delayed edges.

\end{abstract}

\section{Introduction}
Deep learning based on convolution neural networks (CNN) is considered as one of the most successful methods in modern machine learning and computer vision fields. It generally requires a large amount of data samples for real world applications, yet there exist physical limitations and privacy issues in acquiring more and more data.

Federated learning (FL) \cite{yang2019federated, ludwig2020ibm}, a recently suggested concept, is one of the many solutions to the challenge of acquiring vast amount of data without \nj{invasions} of privacy. FL assumes that many distributed devices exist for their own data collection \nj{and the} main goal is the parallel or distributed training of those device-specific datasets by one central model.
Since its proposal with deep neural nets, many works have focused on \sh{averaging the weights} of client models to acquire a centralized server model. 

\begin{figure}[t]
        \centering
            \includegraphics[width=0.48\textwidth]{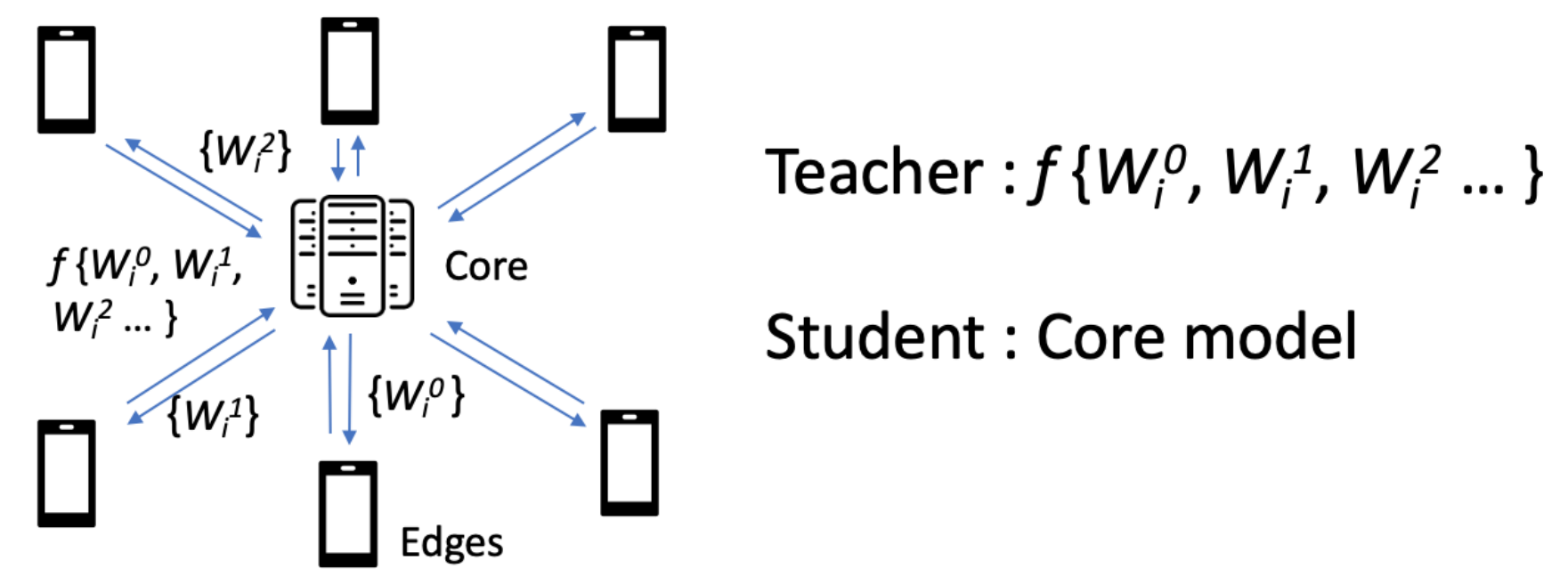}
            \caption{\nj{An illustration of} \textit{knowledge distillation-based} federated learning}
\label{fig:kd_fl}            
\end{figure}

Recently, methods based on Knowledge Distillation~\cite{hinton2015distilling} have shown notable performance \cite{lin2020ensemble, sui2020feded, seo2020federated, chen2020feddistill}  \nj{in the scenario of FL}. Figure \ref{fig:kd_fl} illustrates a general scheme of the Knowledge Distillation-based FL methods. Briefly, these methods aggregate teacher models from the local devices (edge) and trains the server model (core) with the knowledge distillation loss \nj{with respect to the aggregated teachers}. Not only \nj{does the method in} \cite{lin2020ensemble} show higher performance than other methods \cite{Wang2020Federated_matavg}, but it is also highly practical as the method is easily \ky{extendable} to heterogeneous models (e.g. architecture, numerical precision).

In this work, we focus on a particular problem that occurs when utilizing knowledge distillation in FL in certain scenarios, dubbed as ‘edge bias’. We \ky{hypothesize} that this problem ensues from individually distilling knowledge of multiple teacher models trained on different datasets.
That is, in contrary to \nj{the} conventional knowledge distillation scheme, when a teacher model (edge) is either over-fitted to its own dataset or too disparate from the student model (core), the edge model might introduce a \nj{severe} bias in the knowledge distillation process.  

\begin{figure}[t]
\includegraphics[width=0.44\textwidth]{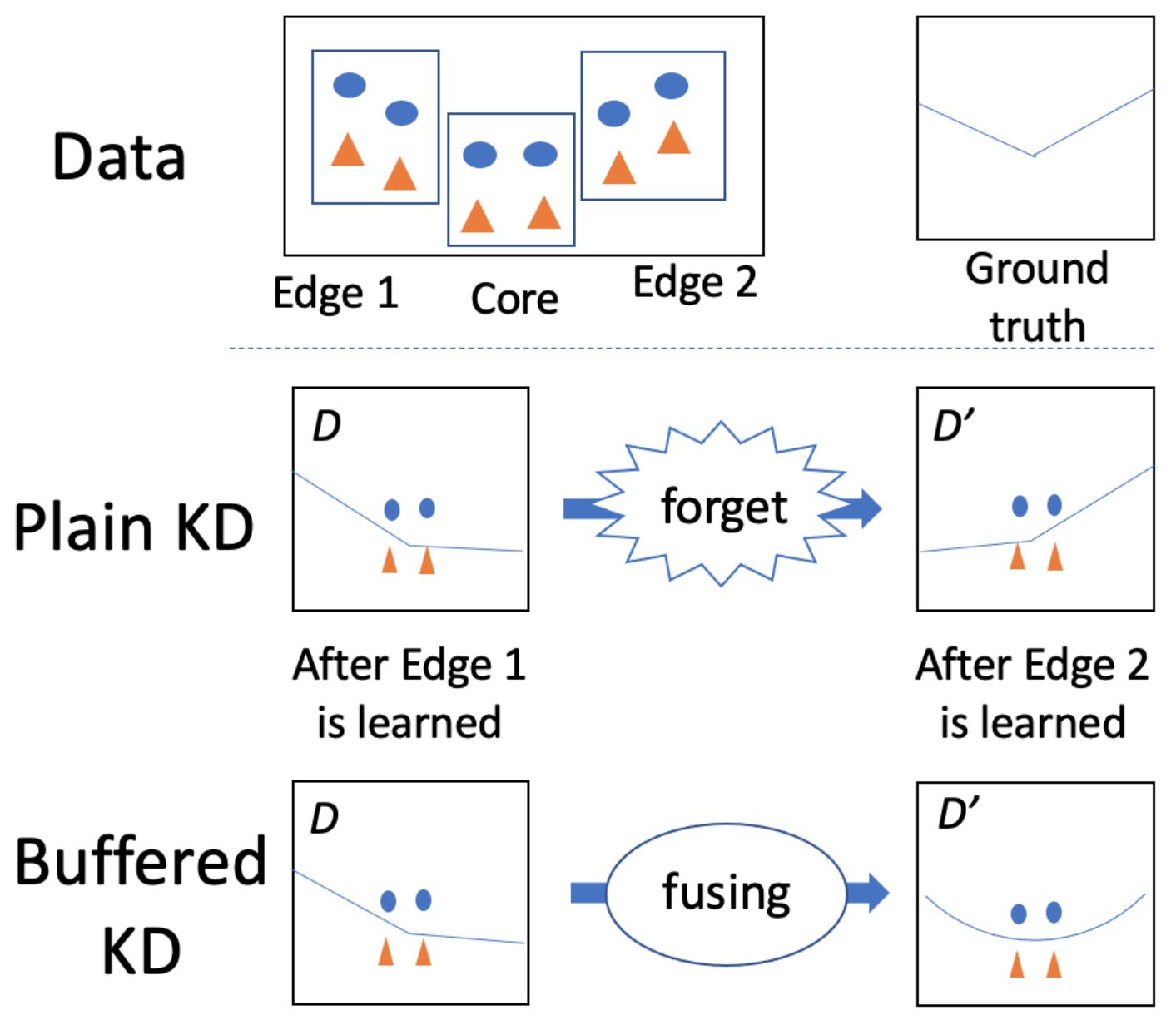}
\caption{The illustration depicting the expected effect of using \textit{buffered Knowledge Distillation} as opposed to vanilla Knowledge Distillation \sh{under \textit{edge bias}. From Edge 1, the ($\backslash$) shaped boundary could be learned by the model, but this knowledge is hardly compatible with \ky{the other dataset} (/), \nj{and} forgetting may happen. But using buffered KD, those \ky{knowledge \nj{from different edges} can be fused with} each other.}}
\label{fig:concept}
\end{figure}

We define `edge bias' as the bias introduced to the teacher model (edge) from training on a particular edge dataset. 
\sh{\ky{Briefly}, \ky{an} edge bias is \nj{the edge's knowledge from its \ky{respective} dataset} that is incompatible with other \nj{edge's knowledge.}} While \nj{a} teacher model \nj{in an edge} may correctly predict on \nj{the corresponding} edge dataset, 1) it may be over-fitted (which is more likely when the edge contains small amounts of data) or 2) it may \sh{induce the student model (core) to \nj{forget its original knowledge} due to} \nj{the disparity between} the two models.    
We expect \nj{our scheme called `\textit{buffered knowledge distillation}’ to} be able to selectively distill knowledge that aligns well with the student model. Figure \ref{fig:concept} depicts the desired effect of our method. We conduct series of experiments to empirically show that our method indeed alleviates this challenging problem \nj{of edge bias}.

As stated above, the edge bias problem is a detrimental consequence of individually distilling biased edge models in repetition. Thus, an apparent solution would be to use an ensemble of all or multitude of edge models for distillation. Indeed, many works in FL including \cite{lin2020ensemble} have focused on model averaging of several edge models to acquire a centralized server model termed as the aggregation phase. However, this phase entails several unrealistic assumptions such as the concurrent response times and training phases of the edge models owned by individual user devices. In practice, aggregation may lead to a bottleneck \nj{causing a network delay with increased communication cost} that linearly increases with the number of aggregated edges at the benefit of performance improvement. 
In this paper, we focus on the scenario of FL with a zero or lightweight aggregation phase. While aggregation leads to a performance improvement in most cases, we open a new line of research that focuses on a more challenging scenario \nj{of distilling knowledge from totally asynchronous edges.} 

In addition, we also show that our method is robust to the straggler problem, one of the continuing challenges of FL, caused by delayed responses of \nj{some} edge models. We conjecture that one explanation of the straggler problem may be due to edge bias. This is further detailed in the experiments section.

All in all, we suggest that the bias of the teacher model may be introduced to the student model in the conventional knowledge distillation scheme for certain scenarios of FL, and we propose a solution called “buffered distillation”. 

Our contributions are as follows:


1) We introduce a problem called edge bias in knowledge distillation-based federated learning methods. \\ 
2) We propose a simple solution called buffered knowledge distillation to tackle \nj{the problem of edge bias}. \\ 
3) We demonstrate that the \nj{proposed} solution is able to not only improve upon federated learning, but also suggest a possible solution to the straggler problem.

\section{Related Works}
With the prevalence of mobile devices that can collect rich variety of data, the concept of training a robust model with decentralized data has emerged as an important research area in both industry and academia (\cite{ludwig2020ibm, Wang2020Federated_matavg, chen2019communication}). 
FL is related to other types of \nj{learning schemes} such as continual learning \cite{kirkpatrick2017overcoming} and semi-supervised learning \cite{berthelot2019mixmatch} in that it utilizes additional streams of data. However, the fact that the central model does not have \nj{a} direct access to the data characterizes FL from other \nj{schemes}. Due to this distinction, simply applying methods from other \nj{schemes} is usually ineffective (e.g. Mean Teacher~\cite{tarvainen2017mean}.)

The first recent proposal in FL \cite{mcmahan2017communication} applied to deep networks used model averaging of the client models to acquire a centralized model.
Since then, many variants of model averaging in the aggregation step have been proposed. \nj{\citet{li2020federated} added} a proximal penalty between the client model and the core model. \nj{\citet{yurochkin2019bayesian} tackled} the permutation invariant property of neural networks that causes a detrimental effect in model averaging by matching neurons of the client models. \nj{\citet{Wang2020Federated_matavg} further extended} this idea to deep neural networks by layer-wise matching and averaging. 

Meanwhile, since \citet{guha2019one} proposed to use Knowledge Distillation for single-round communication, many FL methods have been inspired by the use of Knowledge Distillation \cite{lin2020ensemble, sui2020feded, seo2020federated, chen2020feddistill}. \nj{\citet{lin2020ensemble} approached} the task by using knowledge distillation between the averaged model and the client model. 
Knowledge Distillation (KD) in neural networks was first proposed \nj{in} \cite{hinton2015distilling} to enhance a smaller student network by distilling the knowledge of a larger teacher network via the KL divergence loss between the softmax probability vectors. While first proposed as a model compression method, it has since been widely applied to various tasks such as \nj{the} model stealing \cite{orekondy2019knockoff}.
The effectiveness of KD in such tasks hints at the possibility of utilizing KD \nj{for} federated learning. 
\nj{This line of research constitute the baselines of our knowledge distillation-based federated learning.}


\section{Methodology}
\subsection{Knowledge Distillation-based Federated Learning}
Our work solely focuses on Knowledge Distillation-based FL. As stated, the methodology is highly relevant to \cite{lin2020ensemble}. In this \nj{scenario}, a model is sent as a local learner to each edge device (called the ``edge model"), \nj{each local model learns from its} specific edge dataset, \nj{which} is re-collected to the server, \nj{then acts as} the teacher model for the ``core model" (student).

The re-collected teacher (edge model) performs knowledge distillation with a small, pre-prepared core dataset. Consider the core dataset $C$ whose size is $N_C$, and $K$ edge datasets $\{E_k\}_{k=1}^K$, each \nj{with} the size of $N_{E_{k}}$. For \nj{a sample of input-output pair $(x,y)$}, the softmax probability of the core model is expressed as $F(x)$ and that of the $k$-th edge model is denoted as $f_k(x)$. Let us denote the \nj{number of returned edges by $R$\footnote{We can assume $R=K$ in most cases but at each round due to lost communication and the inherent asynchronous property of edge nodes (e.g. timeout), $R$ can vary and become $R <K$.}} and the ensemble \nj{average of \ky{$\{f_k(x)\}_{k=1}^{R}$}} by $\mathcal{A}_f(x)$. 

Algorithm~\ref{alg:main} describes the overall \nj{process} of KD-based FL.  
The algorithm is composed of 3 phases of which Phase 1 and Phase 2 are iterated. This iteration constitutes a round. In detail, in Phase 0 the core model learns from the core dataset, and each round is executed for each edge that participates in the core service. 

\begin{algorithm}[t]
  \caption{KD-based FL}
 \label{alg:main}
  \begin{algorithmic}
    \STATE {\bfseries Result:} Core model performance
    \STATE Phase 0: Core initialization -- core is trained with the core dataset $C$ with $L_{core}$
    \WHILE{Core model converges}
        \STATE Downlink: Send the core model to the edge: \nj{the edge model becomes identical to the core model}
        \WHILE{edge model has not reached a learning threshold}
        \STATE Phase 1: Learning on edge -- \nj{$i$-th} edge model is trained with the edge dataset $E_i$ with $L_{edge}^i$
        \ENDWHILE
        \STATE Uplink: Send the edge model to the core
        \WHILE{Knowledge distillation has not converged}
        \STATE Phase 2: Knowledge distillation -- transfer edge's knowledge to the core with $L_{KD}$
        \ENDWHILE
    \ENDWHILE
  \end{algorithmic}
\end{algorithm}

In Algorithm~\ref{alg:main}, the loss functions in each phase (Phase 0 through 2) are expressed as follows: 
\begin{equation}
    L_{core} = \sum_{i=1}^{N_C} L(F(x_i), y_i),  
\end{equation}
\begin{equation}
    L_{edge}^k = \sum_{j=1}^{N_{E_k}} L(f_k(x_j), y_j), 
\end{equation}
\begin{equation}
    L_{KD} = L_{core} + \tau^2 \sum_{i=1}^{N_C} KL(F(x_i), \mathcal{A}_f(x_i)/\tau). \label{eq:distill}
\end{equation}
Here, $L(\cdot, \cdot)$ and $KL(\cdot, \cdot)$ are the cross entropy loss and KL divergence respectively, \nj{$\mathcal{A}_f(x)$ is obtained from the aggregated edges $\{f_k(x)\}_{k=1}^{R}$}, and $\tau$ is the temperature. \ky{When a single edge is only used for distillation without any aggregation, $R=1$.} 
Note that $C$ and $E_k$ are fully disjoint sets, and each of those loss functions is used only in the specified phase, i.e, in Phase 0, $L_{core}$ is used, while $L_{edge}$ and $L_{KD}$ are used in Phase 1 and Phase 2 respectively.


\subsection{Buffered Knowledge Distillation}
\label{sec:method}
The main emphasis of this paper is \nj{that} `we want to distill only the unbiased knowledge of the teacher model to the core model, not the biased knowledge', under \nj{the scenario of federated learning}.
Before moving on, let us define the \nj{`biased' and `unbiased knowledge'} \nj{of a model} for the sake of clearness.
First, we define the `knowledge' of a model as the model's implicit factors (e.g. features, weight) that determine the predictions for the data points in the data space. Any machine learning algorithm could make \nj{a} model to learn some knowledge from some finite dataset $D$. \ky{Even \ky{if} the model's predictions on $D$ is largely correct, the predictions of points outside $D$ may deviate from the ground truth, \sh{depending on what knowledge (implicit factors) is learned}.}
From this definition, we suggest that the knowledge of a model can be separated into two categories, which are `biased' and `unbiased' \nj{knowledge}. 

\ky{The `biased knowledge' of a model obtained from $D$ with respect to $D'$ is defined as the implicit factors 
that are incompatible \nj{with} a different dataset $D'$. Conversely, `unbiased knowledge' indicates the equivalent components or compatible factors 
with respect to \nj{the ones from $D'$}. The incompatibility may be due to a large discrepancy of the \nj{two sets of} knowledge, 
\nj{which can be detrimental when one tries to fuse the two (e.g. forgetting or overfitting may occur). }}
In our scenario, $D$ and $D'$ are the dataset for the previous edge and the current edge respectively. The biased knowledge then would be the edge bias of $D'$ with respect to $D$ and vice versa. 


We \ky{conjecture} that edge bias is characterized by two properties. First, a biased edge will have knowledge that can only explain the particular data used in training, but unable to explain unseen data. This form of overfitting will result in an edge teacher model with near 100\% training accuracy, but a low test accuracy. Works such as \cite{zhang2018interpretable}  show that a neural network is able to memorize non-contextual information such as the background to achieve high accuracy in the training data. 

Second, even if the knowledge attained by the teacher model is correct, the knowledge may not be compatible with the student model if the discrepancy \nj{between the two} is too large. Due to the large discrepancy, the student model may only learn the teacher's \nj{new} knowledge at the cost of forgetting its original knowledge.

\nj{
These properties can be more clearly illustrated by Fig. \ref{fig:concept}. In the figure, each edge \sh{data and core data} are \sh{disjoint and} \ky{cover} \ky{respective} portions of the entire data space.  
If we apply conventional KD with the Edge-1 teacher model, (\textbackslash)-shape knowledge is distilled \ky{to the core model}.
\ky{Subsequently,} if KD is \ky{once again} applied with the Edge-2 teacher, (/)-shape knowledge from Edge-2 is distilled which incurs the undesirable forgetting of Edge-1's knowledge\footnote{In a sequential learning setting such as continual learning, it has been shown that overparametrized networks easily forget.}.
In short, two models with incompatible (largely different) knowledge cannot be easily fused in the scenario of sequential knowledge distillation. 
}

\begin{figure}[t]
\includegraphics[width=0.48\textwidth]{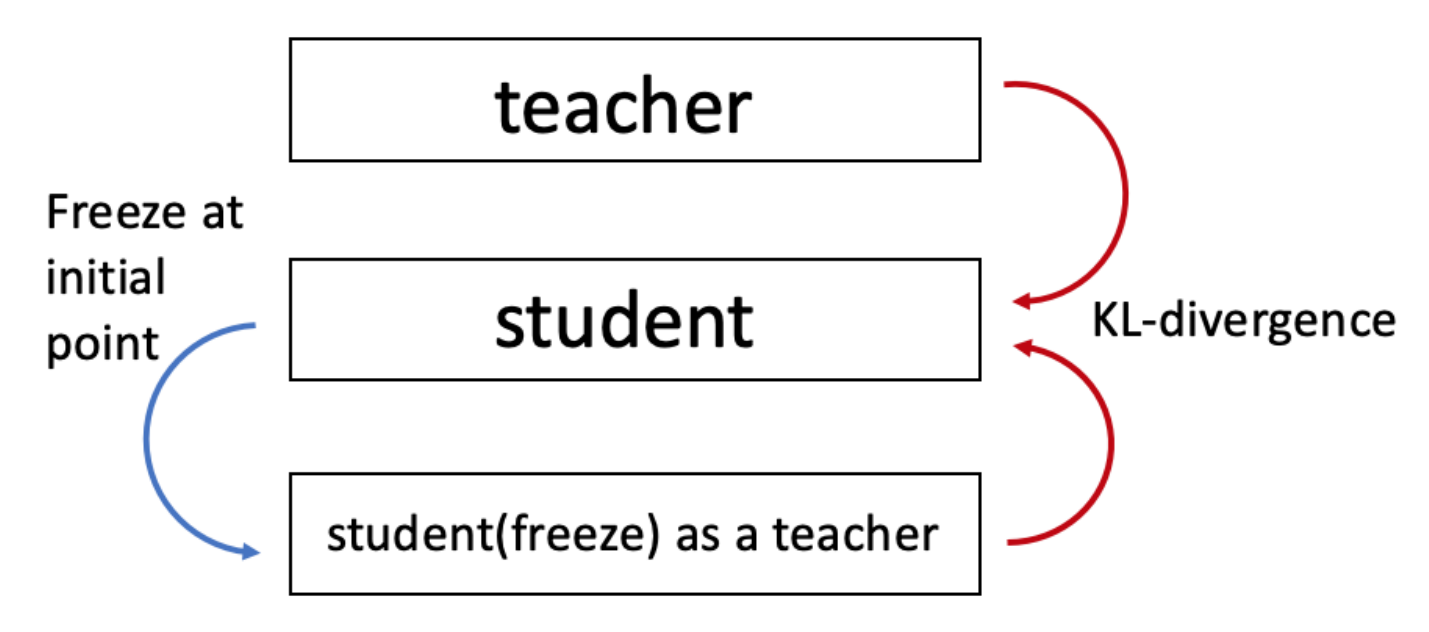}
\caption{A figure describing the training process of buffered KD. A clonded model of the student is freezed \sh{at the start of the training} and used as the \nj{second} teacher for KD.}
\label{fig:BKD}
\end{figure}

\sh{To selectively distill the unbiased and easily fusible knowledge}, the model needs to remember \nj{some of the} previously learned knowledge. Fig. \ref{fig:BKD} shows \nj{a} simple method how \nj{this is done by distilling from two sources: 1)} cloned student \nj{model containing previous knowledge and 2) the teacher with new knowledge.} \ky{The} cloned student is \nj{frozen} at the start of phase 2 training.
\nj{The proposed method is called as buffered KD (BKD) \ky{as the cloned student model is used as a buffer when distilling the new edge's knowledge}, which can be expressed by adding a term to Eq. (\ref{eq:distill}) as follows:}
\begin{equation}
\begin{split}
    L_{BKD} &= L_{KD} + \tau^2 \sum_{i=1}^{N_C} \mathrm{KL}(F(x_i), F_{0}(x_i)/\tau),
    \label{eq:buffered_distill}
    \end{split}
\end{equation}
\nj{where $F_0(\cdot)$ denotes the cloned student model frozen at the start of training.} 

\sh{
If the core \nj{were able to} access the previous dataset, the ideal way would be to train on both the previous and the current datasets at the same time. This would render the model with knowledge unbiased to either one of the dataset\nj{s}. However, because this is not possible in FL, we use KD to distill the knowledge gained from the previous dataset \nj{by} buffering it with a cloned teacher. \nj{With the two sources of teachers, the student model at the core can} distill the more compatible \nj{knowledge (see the last row of Fig. \ref{fig:concept})}.      
}

\section{Experiments}
We followed the overall experimental settings of \cite{lin2020ensemble} and \nj{tested the proposed method} on CIFAR-100~\cite{krizhevsky2009learning} and ImageNet~\cite{imagenet_cvpr09}. For CIFAR-100, the \sh{train} dataset was divided into 20 subsets while for ImageNet, 150 subsets were created, one of which was used as the core dataset. ImageNet was down-sampled to $32 \times 32$ \nj{\ky{to expedite} experiments}. Since the scope of this work lies in \nj{coping with} edge bias, we did not use \nj{any} unlabeled dataset in the core for knowledge distillation like \cite{lin2020ensemble}. To create \sh{non-i.i.d} subsets with different ratios of samples per class, the subsets were sampled using the Dirichlet distribution with $\alpha=1$, \nj{i.e, uniformly sampled from $C-1$ probability simplex where $C$ is the number of classes for each dataset.}

For CIFAR-100, ResNet-32~\cite{he2016deep} was used in \nj{each} edge for 160 epochs with learning rate 1e-1 \nj{which was decayed by a factor of 10 at 80/120 epochs}. For ImageNet, ResNet-32 was used to train locally for 80 epochs starting from \nj{a learning rate of} 1e-1 \nj{which was decayed by a factor of 10 at 40/60 epochs}. \sh{Batch size was set to 128 for both cases. We used \ky{one} NVIDIA TITAN RTX for all experiments. \nj{Experiments on CIFAR-100 took \ky{approximately 5 hours while those on ImageNet took approximately 168 hours.}}} 
For the knowledge distillation loss terms, $\tau$ was all set to 2. In all \nj{the} experiments, we wanted to reduce the effects of hyperparameter tuning, so the weighted sum of each terms in Eq. (\ref{eq:distill}) and (\ref{eq:buffered_distill}) was not considered. Note that the main purpose of our experiment is to verify the effectiveness of buffered distillation \sh{by distilling the unbiased knowledge when heavy aggregation is not possible.}

\begin{figure}[t]
\centering
\begin{subfigure}[b]{0.4\textwidth}
\centering
\includegraphics[width=\textwidth]{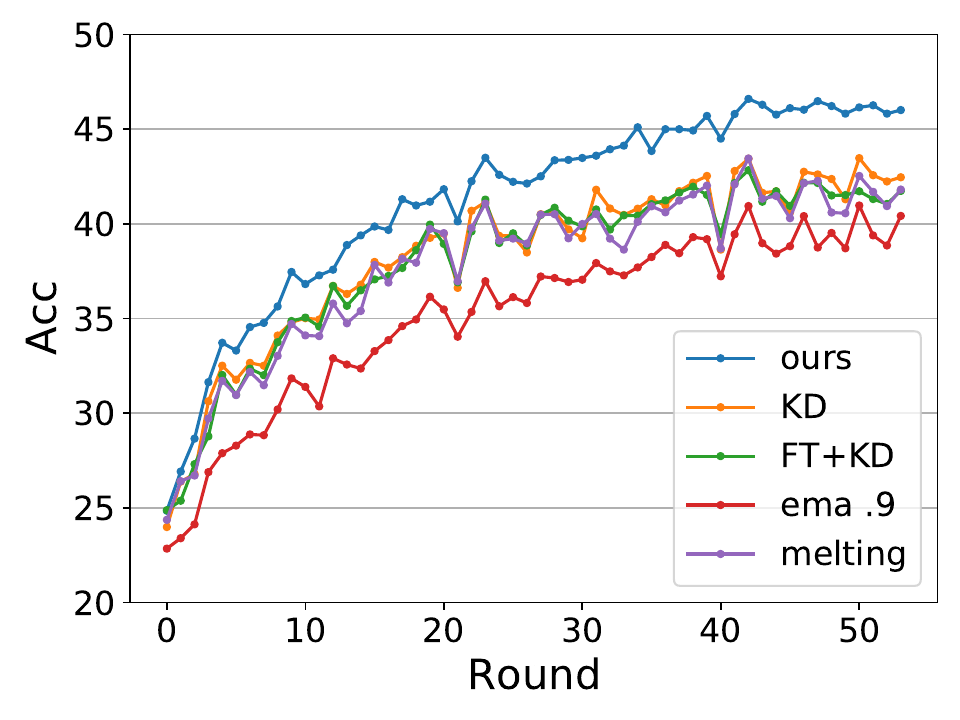}
\caption{CIFAR-100}
\end{subfigure}
\hfill
\begin{subfigure}[b]{0.4\textwidth}
\centering
\includegraphics[width=\textwidth]{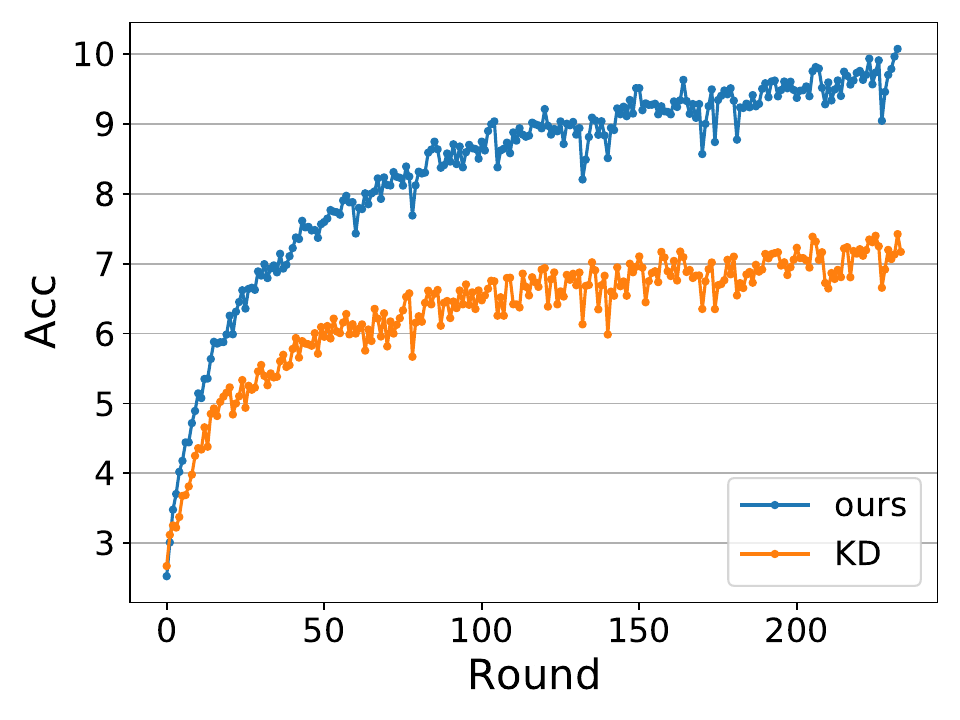}
\caption{ImageNet (resolution 32)}
\end{subfigure}
\caption{Experiment results for CIFAR-100 (19 edges with ResNet-32) and ImageNet (149 edges with ResNet-32) with single edge aggregation $R=1$. The horizontal axis denotes the communication round and the vertical axis denotes the test accuracy of the core model. The dataset of each edge is sampled from the Dirichlet distribution with $\alpha=1$.}
\label{fig:idle_seq}
\end{figure}

\subsection{Main Experiments}
Figure \ref{fig:idle_seq} shows the experimental results of using only a single edge without aggregation \nj{($R=1$)} for \nj{both} CIFAR-100 and ImageNet \nj{datasets}. This is when \nj{the} edge bias is most evident. 
KD \nj{corresponds to} a special case of \cite{lin2020ensemble} when $R=1$\footnote{\ky{The fewest number of edges considered in the original paper were 8 for CIFAR-100 and 15 for ImageNet}}. As shown by the figure, our method achieves higher performance at all rounds and \nj{also} eventually achieves higher final accuracy. 

To show that this is not simply a result of using a better KD method, we also experimented with \nj{Factor Transfer~\cite{kim2018paraphrasing},} one of the SOTA methods in KD, which is denoted by FT+KD. \nj{It} also achieved a similar performance \nj{to that of} the vanilla KD method \nj{(Fig. \ref{fig:idle_seq}(a))}. 
This is also verified when using buffered distillation in a conventional knowledge distillation \nj{scenario}, where the teacher and the student share the same dataset. When \ky{training} ResNet-32 as the teacher and the student \ky{with a pre-trained teacher model}, buffered distillation and vanilla distillation yield\nj{ed  similar accuracy}
\nj{(teacher: 68.85\%, student: 68.81\%, student (KD): 69.33\%, student(BKD): 69.25\%.)}
\ky{The student} of FT+KD \nj{performed} 70.86\%. 
However, in federated learning, edge bias may be \nj{severe} because the teacher and the student models are trained on disjoint data and the teacher is continuously replaced with a different teacher, especially if \nj{a single edge is used as a teacher}. This shows that the performance gain of buffered distillation is not simply \nj{due to} a better KD method, but \nj{\ky{originates} from} mitigating \nj{the} edge bias.  

In Figure \ref{fig:idle_seq}(a), EMA denotes using exponential moving average of the weights when using KD. The decay rate of 0.9 was used and a higher value degraded the performance, while a lower value like 0.09 performed similarly to KD. This shows that simply allowing the model to retain more of the core model’s weight will not solve the problem. To alleviate \nj{the effect of} edge bias, the core model has to selectively receive the edge model’s knowledge, rather than \nj{using} a smoothed update of the weights, which further \nj{degrades} the performance. The comparison of buffered distillation, KD, and EMA shows that our method \nj{does not} simply attain a smoothed version of the student model, but selectively aligns the information \nj{of} the teacher and the student. 

\sh{The \ky{'melting', as opposed to a 'frozen' student} in Figure \ref{fig:idle_seq}(a) denotes \ky{the result of using a second teacher that is cloned} at the start of the every epoch, \ky{instead of at the very start of phase 2 training}. This experiment \ky{once again demonstrates how the initially cloned student is \nj{useful}.} If the knowledge of the \ky{initial student} is not preserved, the accuracy gain disappears and \nj{falls back} to the vanilla KD. \ky{The result verifies that} the accuracy gain is not \nj{merely due from} an effect of smooth updates nor \nj{from a} regularization effect of using a second teacher.}

\begin{figure}[t]
\centering
\begin{subfigure}[t]{0.235\textwidth}
\centering
\includegraphics[width=\textwidth]{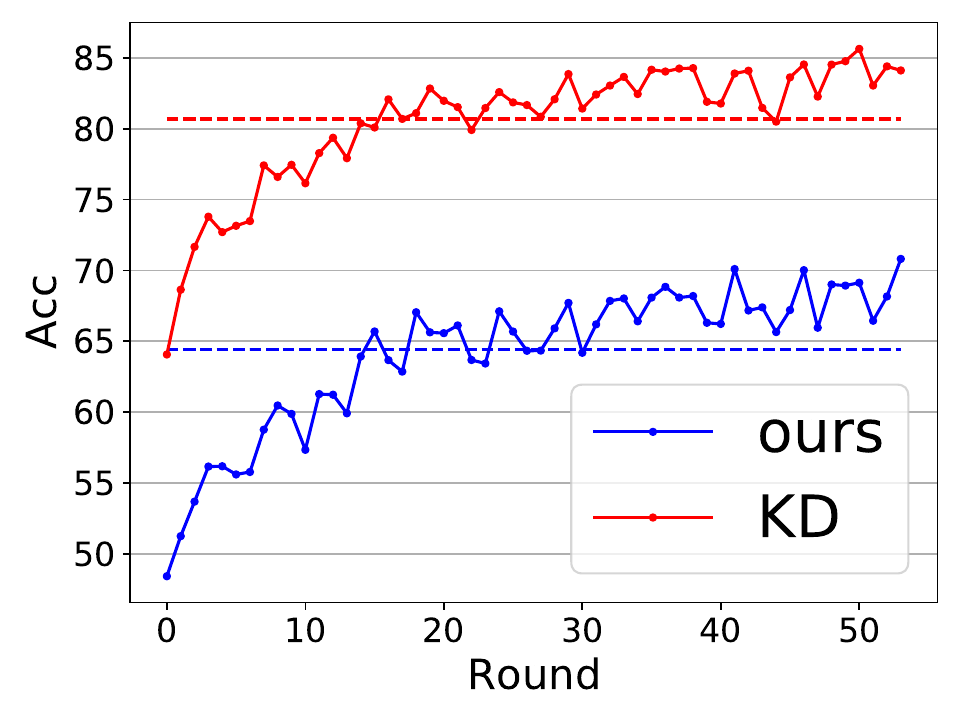}
\caption{Accuracy of core  on $E_t$}
\end{subfigure}
\begin{subfigure}[t]{0.235\textwidth}
\centering
\includegraphics[width=\textwidth]{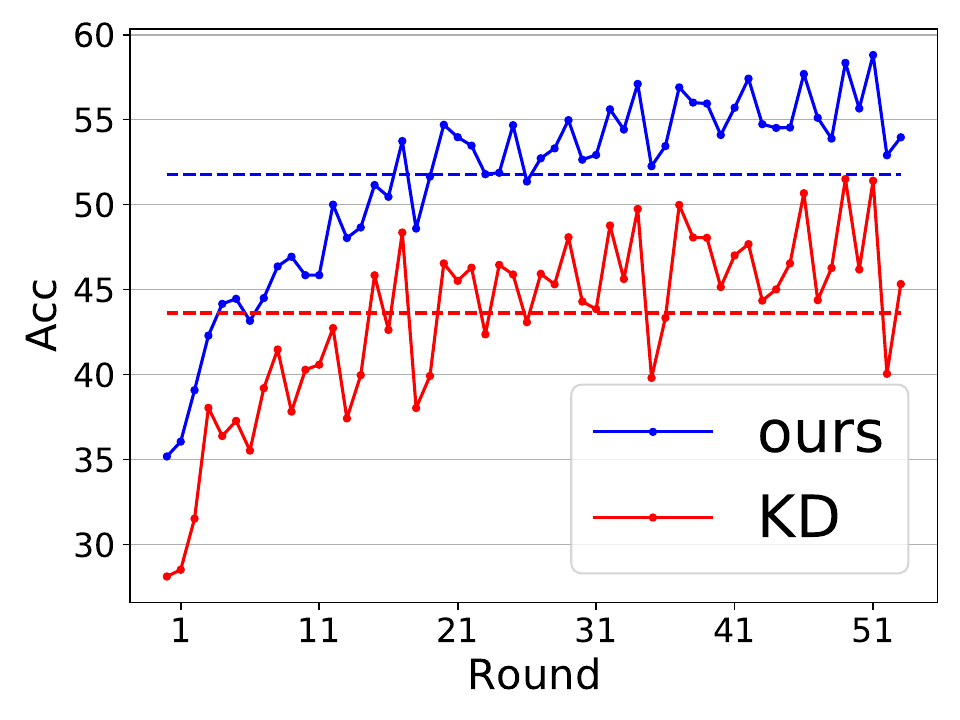}
\caption{Accuracy of core on $E_{t-1}$}
\end{subfigure}
\caption{The figure shows how our method allows the core model to receive the unbiased knowledge by comparing the core model accuracy on the current ($E_t$) edge dataset (a) and the previous ($E_{t-1}$) edge dataset (b). \ky{The dotted lines indicate the mean accuracy over all rounds}. 
}   
\label{fig:mean_forget_score}
\end{figure}

Figure~\ref{fig:mean_forget_score} further illustrates the characteristics of the core model trained with buffered distillation on CIFAR-100. In (a), we show the accuracy of the core model \nj{for the current edge dataset, $E_t$,} after \nj{knowledge distillation has} completed \nj{using the edge model trained on} $E_t$, and in (b) we show the accuracy of the same \nj{core} model on \nj{the previous dataset,} $E_{t-1}$. From (a), we see that core model trained with our \nj{BKD} has lower accuracy than KD. The accuracy of KD on $E_t$ is higher than the test accuracy (Figure \ref{fig:idle_seq}), which shows that the model has overfitted to $E_t$. Note that each edge model in each dataset shows nearly 100\% training accuracy. This shows that \nj{the} core model using buffered distillation is able to not overfit to the edge dataset. In contrast, as shown in (b), our method attains higher performance on the previous edge dataset.
\ky{Note that the drop from the accuracy of $E_t$ (a) to the accuracy of $E_{t-1}$ (b) is larger for KD. This indicates that the biased knowledge of $E_t$ learned by model had a greater effect, leading to forgetting of the previous knowledge. Our model received more of the unbiased knowledge, which is shown by the less impact in accuracy on $E_{t-1}$.}

Subtracting the core model accuracy on $E_{t-1}$ from that of $E_t$ for $t \geq 1$ yields us a quantitative metric of how much forgetting has happened, \sh{that we called mean forget score}, \ky{which is shown in the \nj{supplementary material}}. 

\begin{figure}[t]
\centering
\includegraphics[width=0.4\textwidth]{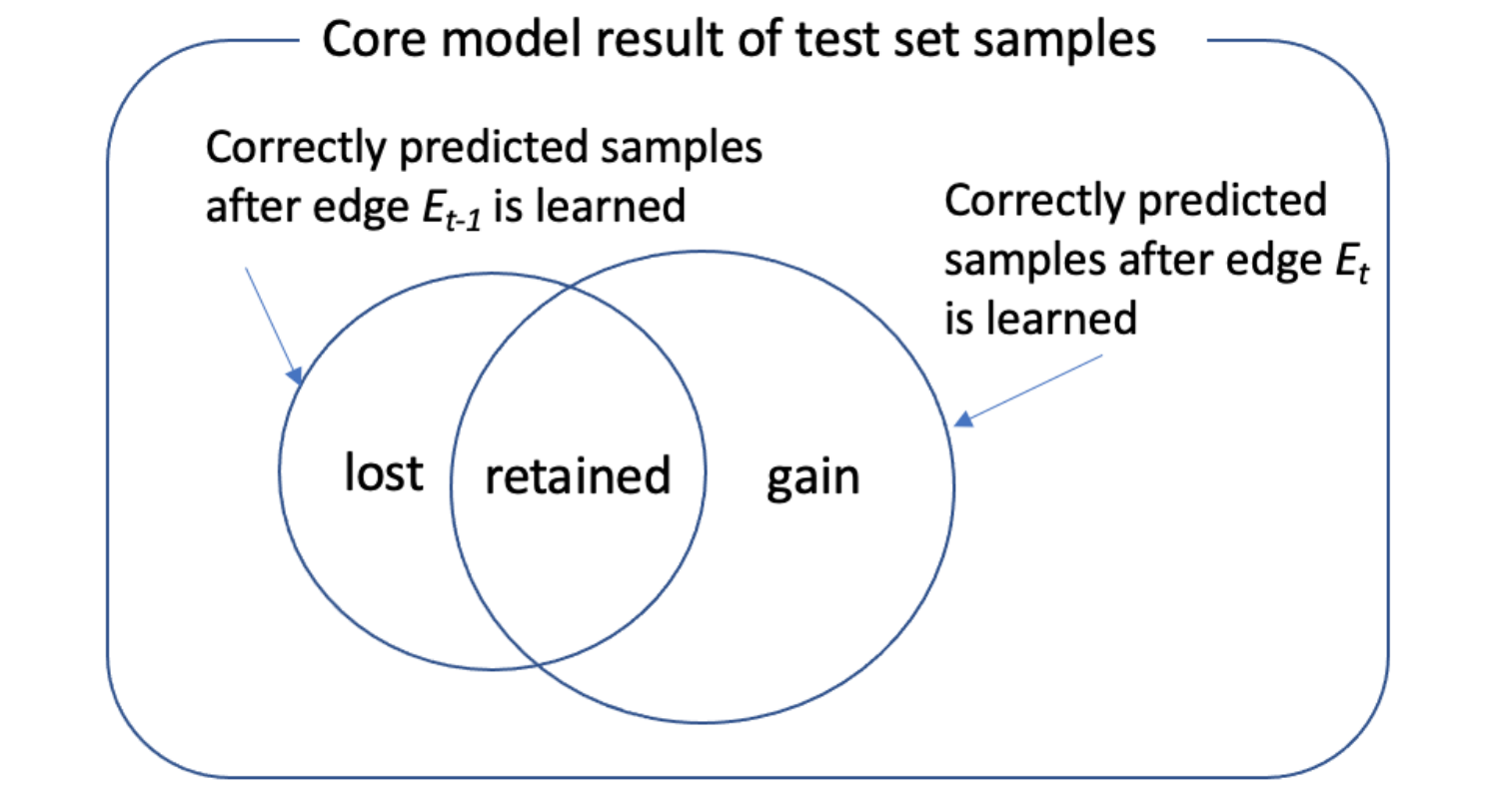}
\begin{tabular}{|c|c|c|}
\hline
                 &  plain kd &  ours     \\\hline
Lost samples     &  1199.4   & 729.5     \\\hline
Gained samples     &  1309.4   & 853.3     \\\hline
Retained samples &  2083.3   & 2728.6    \\\hline
\end{tabular}
\caption{Table showing how the correct prediction on $E_{t-1}$ of the core model changes after training of $E_{t}$ on average for 20 rounds on CIFAR-100. `Lost' signifies the correct predictions that are wrongly predicted after training, `Gain' signifies newly correct predictions, and `Retained' signifies the retained correct predictions.} 
\label{fig:forget_qualitative}
\end{figure}

\begin{figure}[t]
\centering
\begin{subfigure}[t]{0.235\textwidth}
\centering
\includegraphics[width=\textwidth]{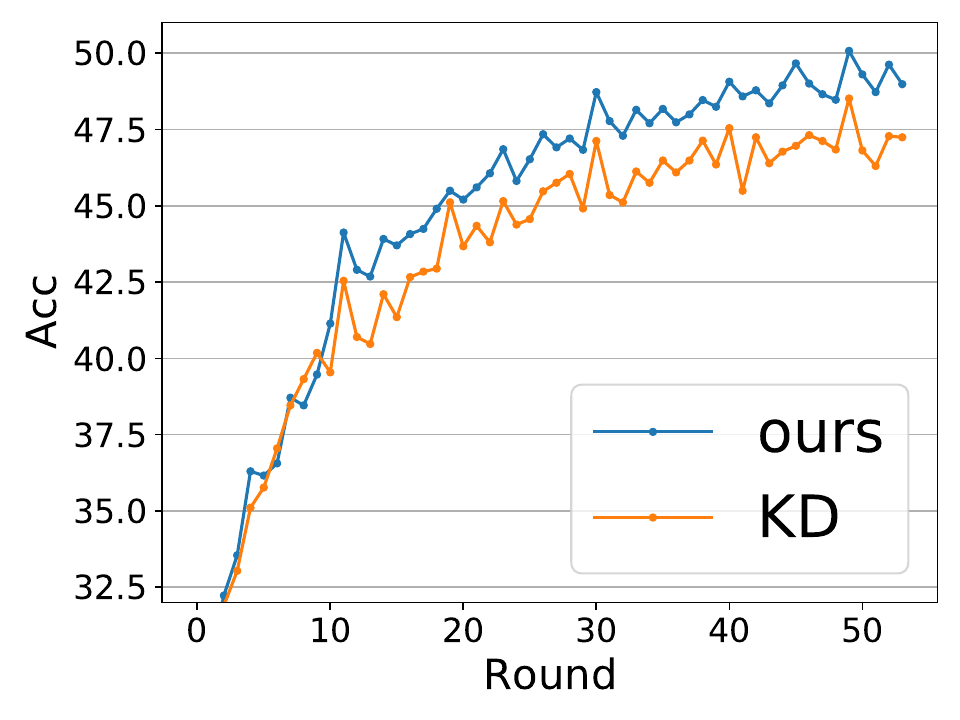}
\caption{CIFAR-100}
\end{subfigure}
\begin{subfigure}[t]{0.235\textwidth}
\centering
\includegraphics[width=\textwidth]{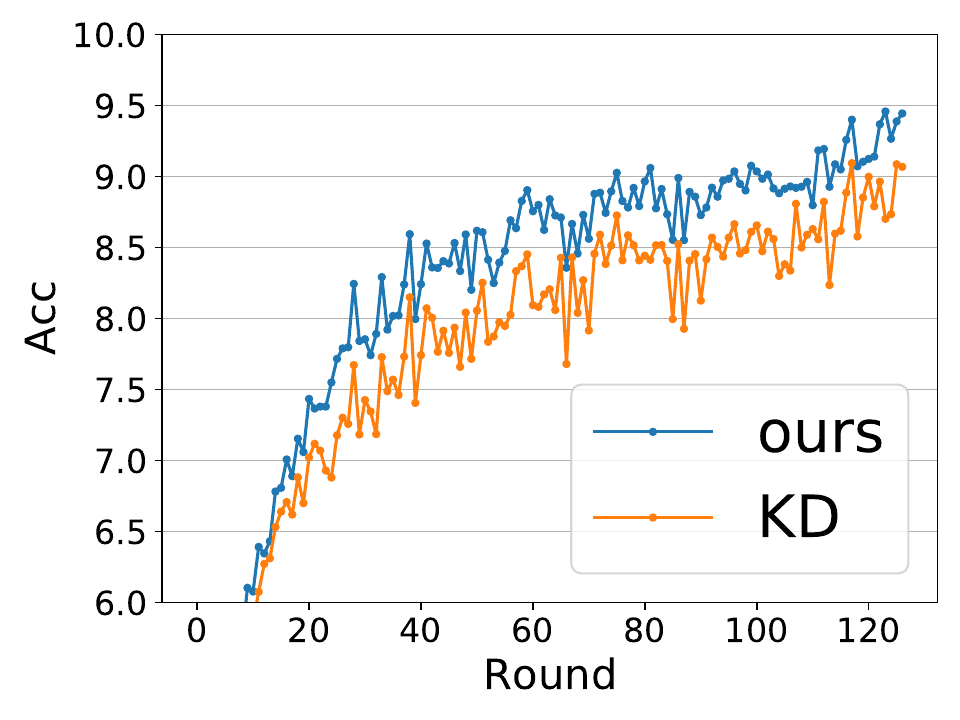}
\caption{ImageNet (resolution 32)}
\end{subfigure}
\caption{Results \nj{of lightweight aggregation with two edges ($R=2$)}. The ensemble of the \nj{two} edge models is used as the teacher.}
\label{fig:agg}
\end{figure}

Figure \ref{fig:forget_qualitative} qualitatively shows the aforementioned effect. This shows how the prediction of the core model at ${t-1}$ changes after training on $E_t$ by displaying the Venn diagram of the correctly predicted samples for 20 rounds. Our method has less `lost' samples and `gained' samples from training on $E_t$, but maintains more `retained' samples \nj{(see the caption of the figure).} This shows that while KD easily drifts by the bias of the current edge dataset, our method is more conservative in adopting the possibly biased knowledge and selectively receives information that is more compatible with the current knowledge.

Then, how different is the knowledge learned with KD from \nj{that with} buffered distillation? To see how the predictions are different \ky{qualitatively} after learning up to edge 19 \sh{(Round 19) of CIFAR-100 \ky{with} 19 edges}, we compare the new correctly predicted sample sets (correct after learned up to edge \ky{19} - correct prediction of initial core model) of KD, FT+KD, \ky{and} BKD. We observe the intersection ($\cap$) over union ($\cup)$ of those sample sets between KD vs BKD is lower than KD vs FT+KD (51.4\% vs 53.0\%) and \ky{BKD vs. FT+KD is 50.0\%}, which \ky{shows that the newly learned knowledge of BKD is more disparate with KD as KD is to FT+KD.}

\subsection{Lightweight Aggregation}
As noted above, our method is most effective when edges \nj{asynchronously} arrive \nj{and distillation is done upon arrival of an edge} as distilling single edge poses the most bias. \nj{A slower distillation scheduler can aggregate several edges and} 
mean ensemble can alleviate \nj{quite an amount of edge bias}. 
\nj{However, the trade-off between reduced edge bias and increased latency should be considered and a plausible scenario would be aggregating a few edges at a time.}
In this section, we empirically show that buffered distillation can also be useful when \nj{this} lightweight aggregation is possible. 

We experimented in the lightweight aggregation scenario \nj{with $R=2$}. \sh{This case of KD \nj{corresponds to} a lightweight version of \cite{lin2020ensemble}}. We observed that the learning curve of buffered distillation increases too gradually for the first few rounds. Also, we observed that in the early training phase of using vanilla KD, the mean forget score follows our trend as shown in Figure \ref{fig:mean_forget_score}. For instance, for the first 4 rounds of communication in CIFAR-100, the accuracy of the core model in $E_{t-1}$ and $E_t$ is on average  \{35.0, 47.5\} of KD, while ours similarly have \{34.8, 42.7\}. Thus, for $R > 1$, we used KD for the first few rounds, then use buffered distillation afterwards, which turns out to be more effective. Figure \ref{fig:agg} shows that \nj{for both CIFAR-100 and ImageNet, higher accuracy is obtained} by using buffered distillation.


\begin{figure}[t]
\centering
\begin{subfigure}[b]{0.5\textwidth}
\centering
\includegraphics[width=0.8\textwidth]{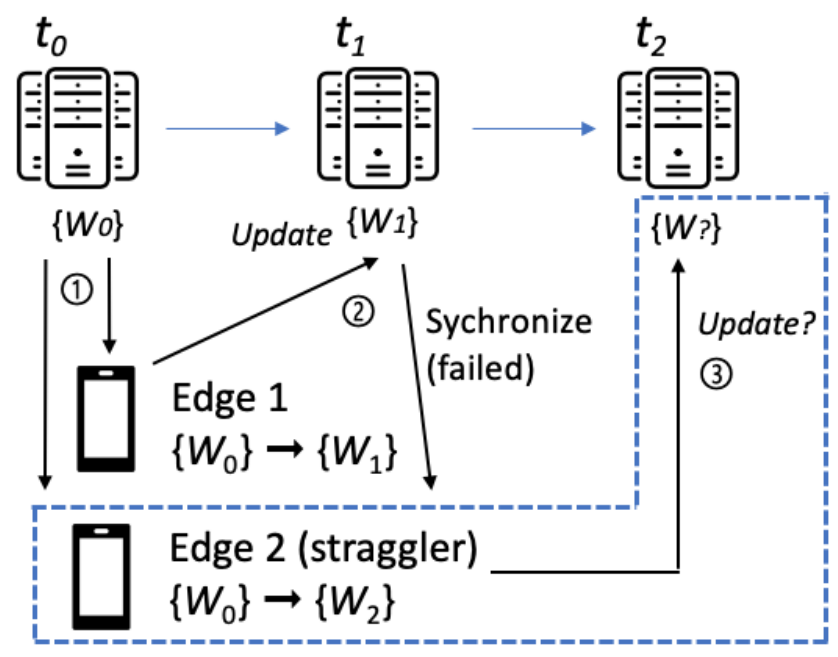} 
\caption{Straggler Edge 2}
\end{subfigure}
\begin{subfigure}[b]{0.5\textwidth}
\centering
\includegraphics[width=0.8\textwidth]{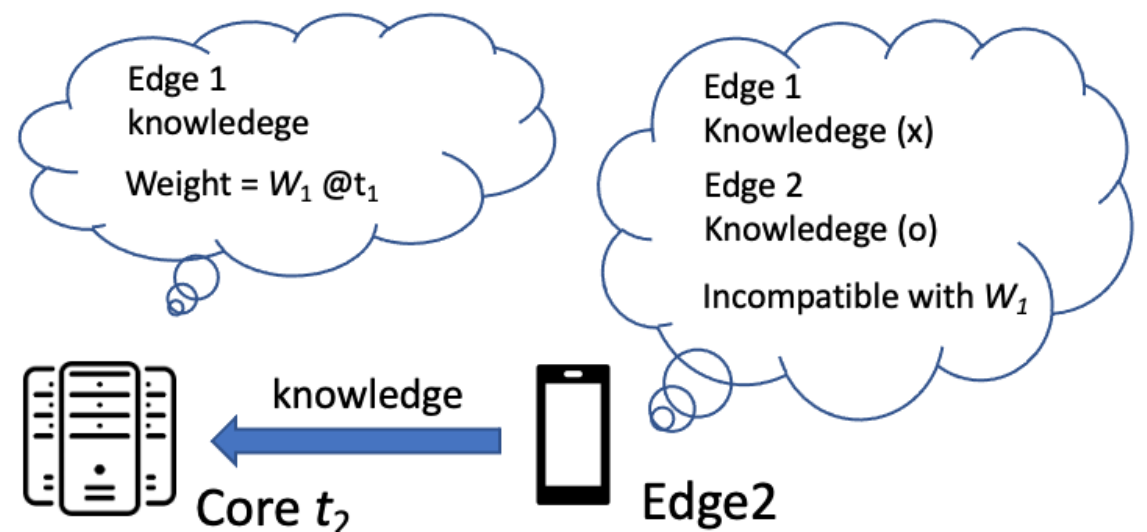}    
\caption{Information asymmetry between Edge2 and Core \nj{@$t_2$}}
\end{subfigure}
\caption{Diagram illustrating the straggler problem handled in this section. In (a), After Edge 1 is updated yielding the core model $W_1$, Edge 2 \nj{is} unable to train with the updated core model and instead \nj{uses} $W_0$. The updated Edge 2 is a straggler, which has information asymmetry with the core and it might have a detrimental effect on the core as shown in (b). 
} 
\label{fig:strag_problem}
\end{figure}

\begin{figure}[t]
\centering
    \includegraphics[width=0.35\textwidth]{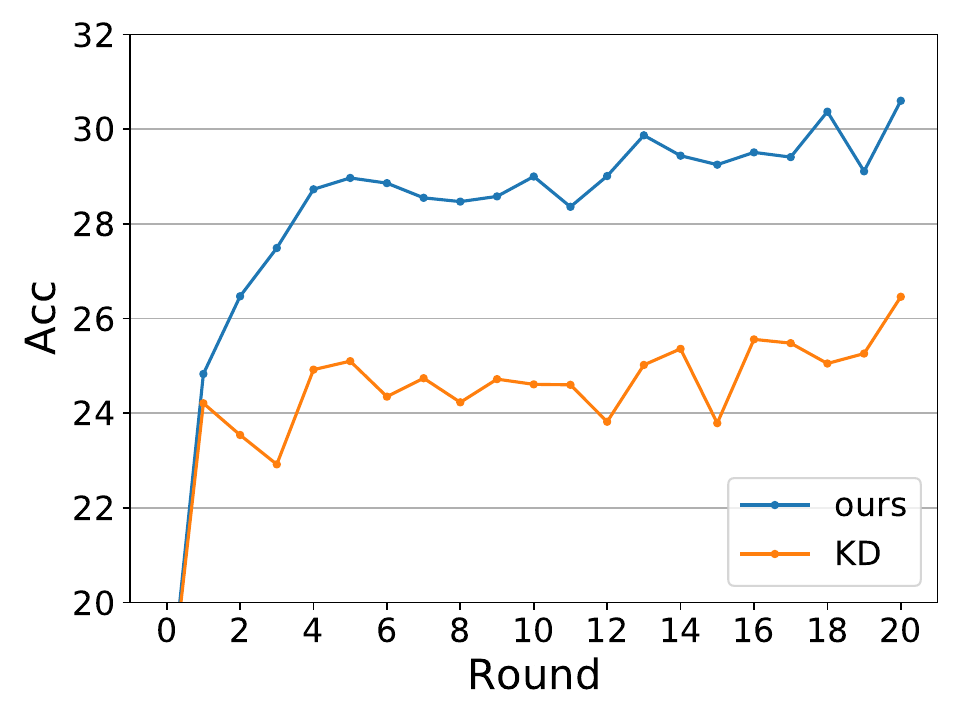}
    \caption{Results on the most extreme case \sh{\nj{of the} straggler problem}, in which all edges are not synchronized after receiving the first core model, $W_0$. Contrary to the one trained with KD, the accuracy of \nj{our BKD} steadily \nj{increases}.}
    \label{fig:long_range}
\end{figure}

\begin{figure*}[t]
    \centering
    \includegraphics[width=0.8\textwidth]{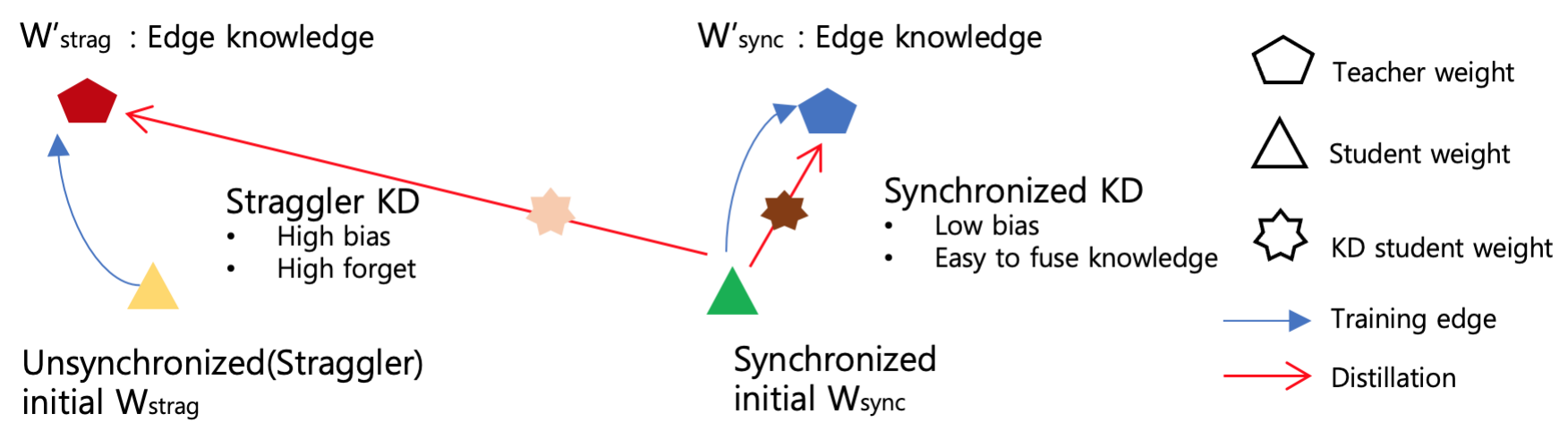}
    \caption{Intuitive interpretation of the difference between unsynchronized KD (straggler) and synchronized KD on an imaginary 2-dimensional parameter space. The properties of the two types of distillation are \nj{described in the text}.}
    \label{fig:stars}
\end{figure*}

\subsection{Straggler Problem}
\label{sec:straggler_problem}
\sh{In this section we deal with the straggler problem \cite{li2014communication, tandon2017gradient}. The experiments in this section \nj{is on CIFAR-100 with 19 edges. ResNet-32 is used as both core and edge models.}}  In this paper, we define the straggler problem as depicted in Figure \ref{fig:strag_problem}, which \nj{is} mainly caused by \nj{edge models that are out of sync with the} core model, because of \nj{slow learning, network delay}, etc.

\textbf{Importance of Weight Synchronization}
Following the convention of other methods, we assumed in the aforementioned experiments that all edge models are constantly synchronized with the core model. That is, after each update of the core model, all edges receive the updated core model, then each edge trains on its own dataset. However, for a system with several thousands of edges, synchronizing all edges with the latest core model is extremely difficult, leading to some edge models that are trained with the obsolete core model. Figure \ref{fig:strag_problem} depicts this situation, which is termed as the straggler problem in literature. We show that this problem of FL is relevant to edge bias.

Before discussing the matter in detail, we first comment on why synchronization between the core model and the edge models is crucial. Let us assume the extreme case, in which the edge models are not synchronized at all and the first weight of the core model $W_0$ is used throughout the whole process of FL. In this setting, we observed that when the vanilla KD is used, the accuracy of the core model does not steadily increase as shown in Figure \ref{fig:long_range}. \nj{Unlike KD, our buffered KD makes the core model perform continuously better.} This implies the feasibility of using our \nj{BKD} in the extreme case where synchronization is rarely possible.

We believe the first reason behind the degradation in performance is due to the large distance between the initial weights of the edge model and the updated core model, which aggravates the edge bias problem. 
Figure \ref{fig:stars} \nj{illustrates} the difference between \nj{the} synchronized KD and \nj{the} unsynchronized (straggler) KD. In the parameter space, consider the two initial \nj{weights} of the teacher model $W_{strag}$ and $W_{sync}$ for \nj{the} straggler and \nj{the} synchronized teacher respectively. \sh{$W_{strag}$, \nj{which} is \nj{an} initial weight vector of \nj{an out-of-sync edge model,} has some distance with $W_{sync}$.} After learning \nj{each} edge data, the final teacher model is denoted by \nj{a} pentagon ($W'_{strag}$ and $W'_{sync}$) in the figure. 
In the distillation stage \sh{of \nj{the} straggler}, the distance between the student ($W_{sync}$) and the teacher for \nj{the} straggler KD ($W'_{strag}$) is typically so large that the final student's weight after KD will be located somewhere in the middle between $W_{sync}$ and $W'_{strag}$. On the other hand, in synchronized KD, the teacher-student distance ($||W'_{sync}-W_{sync}||/||W_{sync}||$) is \sh{smaller than $||W'_{strag}-W_{sync}||/||W_{sync}||$} \sh{(e.g. 0.38 vs 0.44 for CIFAR-100, 3 subsets, 2 edges)} and the resultant student's weight after synchronized KD will be closer to the original student's weight ($W_{sync}$). This means that the student did not forget about the original knowledge, while also absorbing teacher's new knowledge on the edge dataset. 

The second cause lies in ‘information asymmetry’ between the straggler edge and the current core model. As shown in Figure \ref{fig:strag_problem}\sh{(b)}, the core model after learning from Edge 1 has the information of $E_1$. In contrast, because the Edge 2 model has not been synchronized, it does not have any information of $E_1$. Thus, simply imitating the Edge 2 teacher model may lead to forgetting the $E_1$ information. \sh{Also, the core model (student) is already updated to $W_1$ but \nj{the} straggler edge model (teacher) is still dependent on $W_0$, \nj{without any} information of $W_1$. } 

\nj{Therefore, removing stragglers by frequent synchronization is highly desirable in the FL.}
However, for a large system with numerous edges, synchronizing all edges every time the core model is udpated is almost \nj{impossible}. Thus, preventing the straggler problem is an extremely difficult one.


\begin{figure}[t]
\centering
    \includegraphics[width=0.48\textwidth]{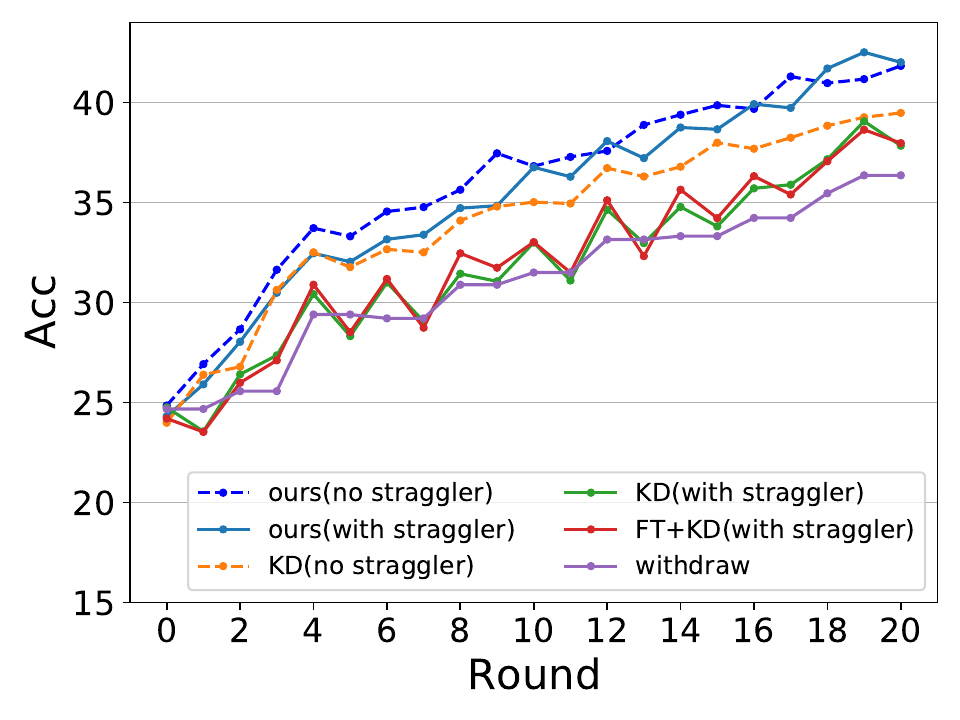}
    \caption{Results on the experiment with straggler edges and synchronized edges alternately used as the teacher model. At each round, the straggler edge / synchronized edge comes in alternately to the core model for distillation. The accuracy of KD fluctuates, while our BKD is relatively robust to stragglers.}   
\label{fig:straggler_sequential}
\end{figure}

\textbf{Robustness to Stragglers}
We show that using buffered distillation to train the core model lends a more robust model to the straggler problem. From the previous experiments, we expect that buffered distillation is less vulnerable to a highly biased or information-asymmetric teacher \sh{or} \nj{a teacher with large discrepancy with the student} as it selectively distills unbiased knowledge.  Fig. \ref{fig:straggler_sequential} shows the results for this experiment following the scenario shown in Fig. \ref{fig:strag_problem}. \ky{To implement the straggler scenario, at every round the edge model alternates between a normal, synchronized model and a straggler model. At the first round, a synchronized model is given and then a straggler is given in the next round. A straggler model at $t+1$ round, for instance, is a model locally trained using $W_t$ instead of $W_{t+1}$.}
The accuracy of KD fluctuates every time a straggler edge is used as the teacher. The ‘withdraw’ denotes the trivial solution of not using the straggler edge, which leads to a lower final accuracy. In contrast, using buffered distillation the fluctuation is greatly mitigated and shows a similar accuracy to that of using buffered distillation without any stragglers. This validates that our BKD yields a model robust to the straggler problem. 


\section{Conclusion}

In this paper, we introduced a potential problem \nj{in} knowledge distillation-based FL, \sh{\ky{called} edge bias. \nj{Edge bias is due from the discrepancy between local datasets in different edges and it causes the core model to overfit to the new edge dataset and forget already learned knowledge from previous edges when a new edge knowledge is distilled. }} 
We proposed a simple solution called buffered distillation to handle this problem, \sh{by selectively distilling the unbiased knowledge of the teacher.} In addition, we provide intuition on why synchronization between the core and the edges is essential for stable performance improvement.  
\nj{Lastly, we applied the proposed buffered distillation to the straggler problem where some stragglers with out-of-date information of the server weights stymie the learning process and showed the feasibility of buffered distillation for this problem.}

\bibliography{example_paper}
\bibliographystyle{icml2021}

\end{document}


\twocolumn[
\icmltitle{Edge Bias in Federated Learning and its Solution \\ by Buffered Knowledge Distillation : Appendix}


\icmlsetsymbol{equal}{*}

\begin{icmlauthorlist}
\end{icmlauthorlist}



\icmlkeywords{Machine Learning, ICML}

\vskip 0.3in
]

\section{Additional Information for \nj{Reproduction}}
For all experiments, SGD with momentum of 0.9 was used for optimization and \nj{the weight decay with a initial learning rate} of 1e-4 was used. For the CIFAR-100 dataset, we used horizontal flip and random crop with padding 4 as augmentations and the images were normalized with the mean (0.507, 0.487, 0.441) and standard deviation (0.267, 0.256, 0.276). For ImageNet, only horizontal flip was used. The inputs were normalized with the mean (0.485, 0.456, 0.406) and standard deviation (0.229, 0.224, 0.225). The downsampling layer (channel expansion block) of ResNet used projection (option `b' in \cite{he2016deep}).

\section{Additional Figure}
\begin{figure}[t]
\centering
\begin{subfigure}[t]{0.32\textwidth}
\includegraphics[width=\textwidth]{fig_mean_after.pdf}
\caption{Accuracy of the core model on $E_t$}
\end{subfigure}
\hfill
\begin{subfigure}[t]{0.32\textwidth}
\includegraphics[width=\textwidth]{fig_mean_before.pdf}
\caption{Accuracy of the core model on $E_{t-1}$}
\end{subfigure}
\hfill
\begin{subfigure}[t]{0.32\textwidth}
\includegraphics[width=\textwidth]{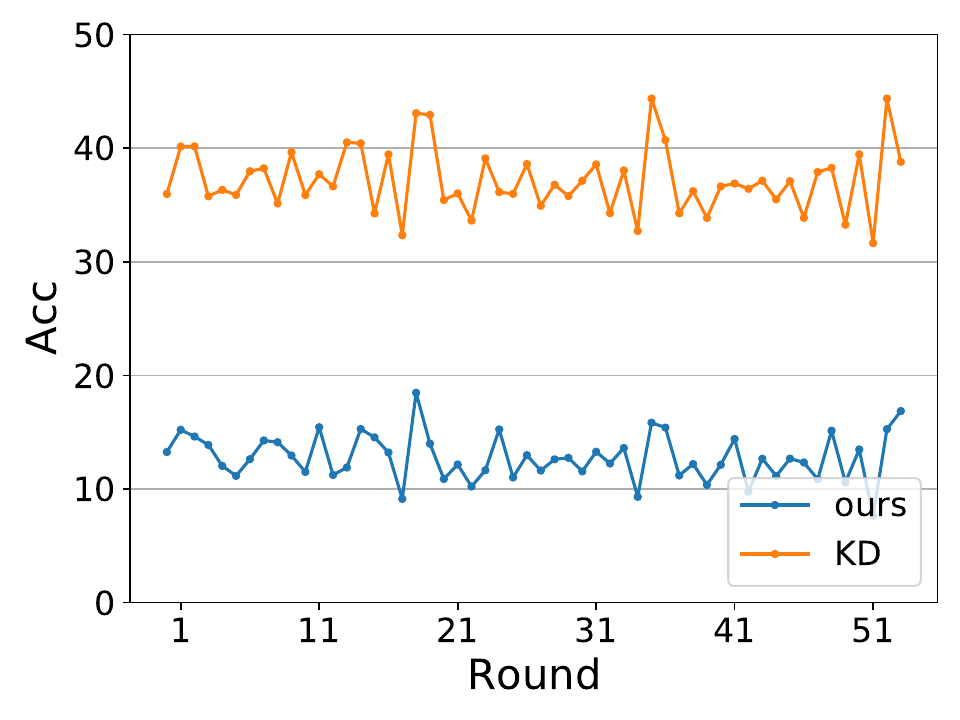}
\caption{Mean forget score (accuracy gap between (a) and (b))}
\end{subfigure}

\caption{The figure shows how our method allows the core model to receive the unbiased knowledge by comparing the core model accuracy on the current ($E_t$) edge dataset (a) and the previous ($E_{t-1}$) edge dataset (b). The dotted lines indicate the mean accuracy over all rounds. (c) shows the difference of the two accuracy.
}   
\label{fig:mean_forget_score_appendix}
\end{figure}

Figure \ref{fig:mean_forget_score_appendix} is the figure that \nj{was} mentioned in Section 4.1.

\bibliography{example_paper}
\bibliographystyle{icml2021}